\documentclass[conference]{IEEEtran}
\IEEEoverridecommandlockouts
\usepackage{cite}
\usepackage{amsmath,amssymb,amsfonts}
\usepackage{algorithmic}
\usepackage{graphicx}
\usepackage{textcomp}
\usepackage{xcolor}
\usepackage{booktabs}
\usepackage{tabularx}
\usepackage{tikz}
\usepackage{pifont}
\newcommand{\cmark}{\text{\ding{51}}}
\newcommand{\xmark}{\text{\ding{55}}}
\newcommand{\etal}{\textit{et al}.}
\newcommand{\ie}{\textit{i}.\textit{e}.}
\newcommand{\eg}{\textit{e}.\textit{g}.}
\newcommand{\etc}{\textit{etc}.}
\newcommand{\vect}[1]{\boldsymbol{#1}}
\newcommand{\methodname}{{\tt{MMLQ}}}

\def\BibTeX{{\rm B\kern-.05em{\sc i\kern-.025em b}\kern-.08em
    T\kern-.1667em\lower.7ex\hbox{E}\kern-.125emX}}
\begin{document}

\title{Multi-modal Learnable Queries for Image Aesthetics Assessment\\
\thanks{This research/project is supported, in part, by the National Research Foundation, Singapore and DSO National Laboratories under the AI Singapore Programme (AISG Award No: AISG2-RP-2020-019); Alibaba Group through Alibaba Innovative Research (AIR) Program and Alibaba-NTU Singapore Joint Research Institute (JRI) (Alibaba-NTU-AIR2019B1), Nanyang Technological University, Singapore; the RIE 2020 Advanced Manufacturing and Engineering (AME) Programmatic Fund (No. A20G8b0102), Singapore.}
}

\author{\IEEEauthorblockN{Zhiwei Xiong$^{1,2}$, Yunfan Zhang$^{1,2}$, Zhiqi Shen$^1$, Peiran Ren$^3$ and Han Yu$^1$}
\IEEEauthorblockA{\textit{$^1$College of Computing and Data Science, Nanyang Technological University, Singapore} \\
\textit{$^2$Alibaba-NTU Singapore Joint Research Institute, Nanyang Technological University, Singapore} \\
\textit{$^3$Alibaba Group, Hangzhou, China} \\
\{zhiwei002, yunfan.zhang, zqshen, han.yu\}@ntu.edu.sg \\ peiran.rpr@alibaba-inc.com}
}

\maketitle

\begin{abstract}
Image aesthetics assessment (IAA) is attracting wide interest with the prevalence of social media. The problem is challenging due to its subjective and ambiguous nature. Instead of directly extracting aesthetic features solely from the image, user comments associated with an image could potentially provide complementary knowledge that is useful for IAA. With existing large-scale pre-trained models demonstrating strong capabilities in extracting high-quality transferable visual and textual features, learnable queries are shown to be effective in extracting useful features from the pre-trained visual features. Therefore, in this paper, we propose \methodname{}, which utilizes \underline{m}ulti-\underline{m}odal \underline{l}earnable \underline{q}ueries to extract aesthetics-related features from multi-modal pre-trained features. Extensive experimental results demonstrate that \methodname{} achieves new state-of-the-art performance on multi-modal IAA, beating previous methods by 7.7\% and 8.3\% in terms of SRCC and PLCC, respectively.
\end{abstract}


\section{Introduction}
Image aesthetics assessment (IAA) refers to evaluating the quality of images based on human perception. With the prevalence of social media, IAA has recently attracted much attention, especially in downstream applications such as image aesthetic enhancement~\cite{li2023all}, and aesthetic-aware text-to-image generation~\cite{xu2023imagereward}. Due to the subjective and ambiguous nature of the definition of aesthetics, the ``standards" of image aesthetics are usually determined by opinion scores from different reviewers. Based on these opinion scores, IAA can be treated as a data-driven problem where the ground truth is either the mean opinion score (MOS) or the distribution of opinion scores (DOS).

\begin{figure}[!t]
\centering
\includegraphics[width=1\linewidth]{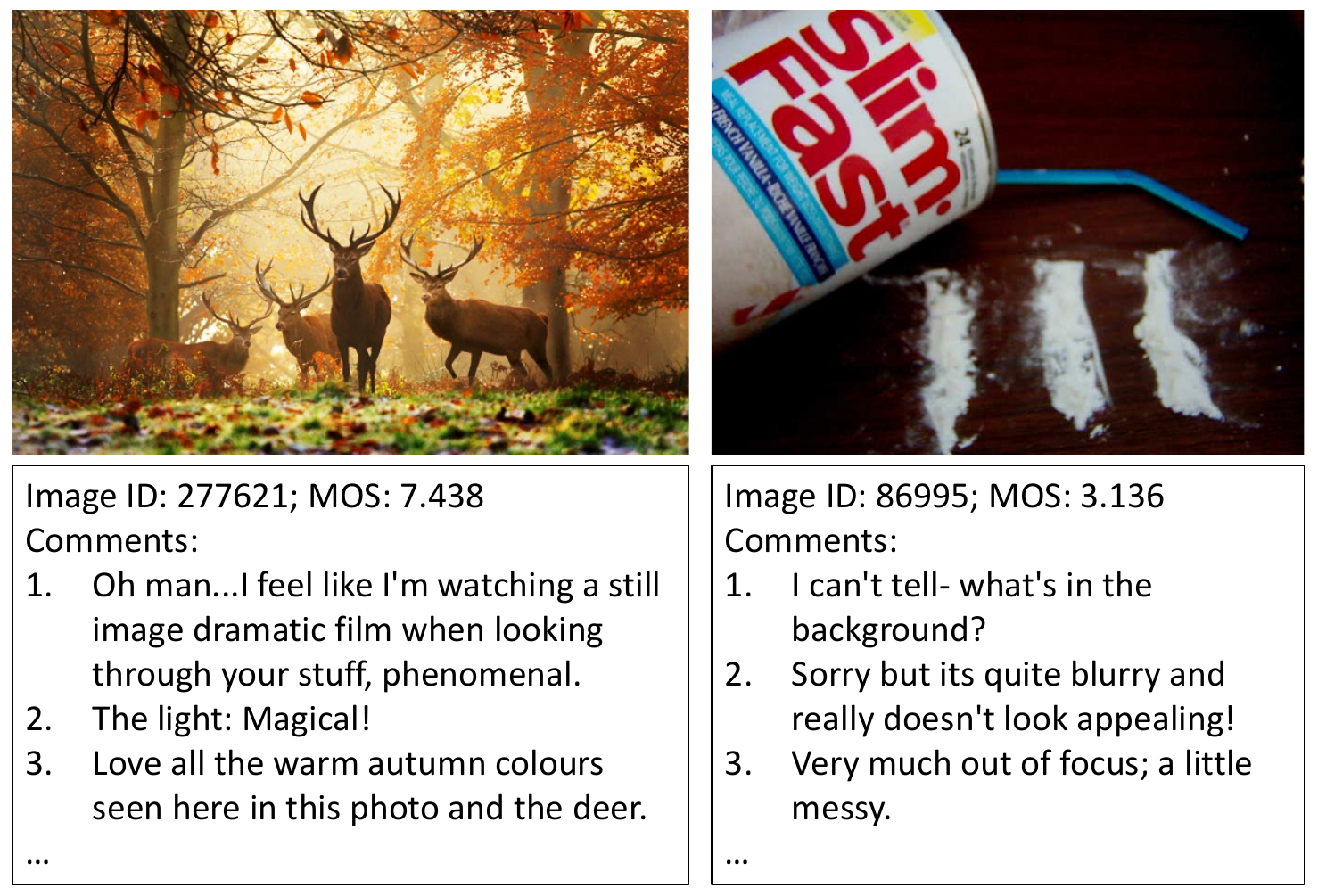}
\caption{Example images with their corresponding image ID, ground truth MOS, and selected user comments from the AVA~\cite{murray2012ava} and AVA-Comments~\cite{zhou2016joint} datasets.} 
\label{fig:ava_samples}
\end{figure}

With direct supervision solely from MOS or DOS, various IAA methods were proposed to extract local to global~\cite{lu2014rapid}, low to high-level features~\cite{hosu2019effective}, and address layout~\cite{she2021hierarchical} to theme-aware~\cite{he2022rethinking} image aesthetics. However, direct supervision from MOS or DOS may suffer from a lack of details regarding how these decisions are made. User comments associated with the images could be utilized to provide additional aesthetic guidance. Fig.~\ref{fig:ava_samples} shows two example images and some of their corresponding user comments from the AVA~\cite{murray2012ava} and AVA-Comments~\cite{zhou2016joint} datasets, respectively. Comments keywords such as ``phenomenal", ``magical", and ``love" for the left image and ``blurry", ``out of focus", and ``messy" for the right image express strong inherent sentiments that could be potentially beneficial for IAA.

\begin{figure*}[!t]
\centering
\includegraphics[width=1\linewidth]{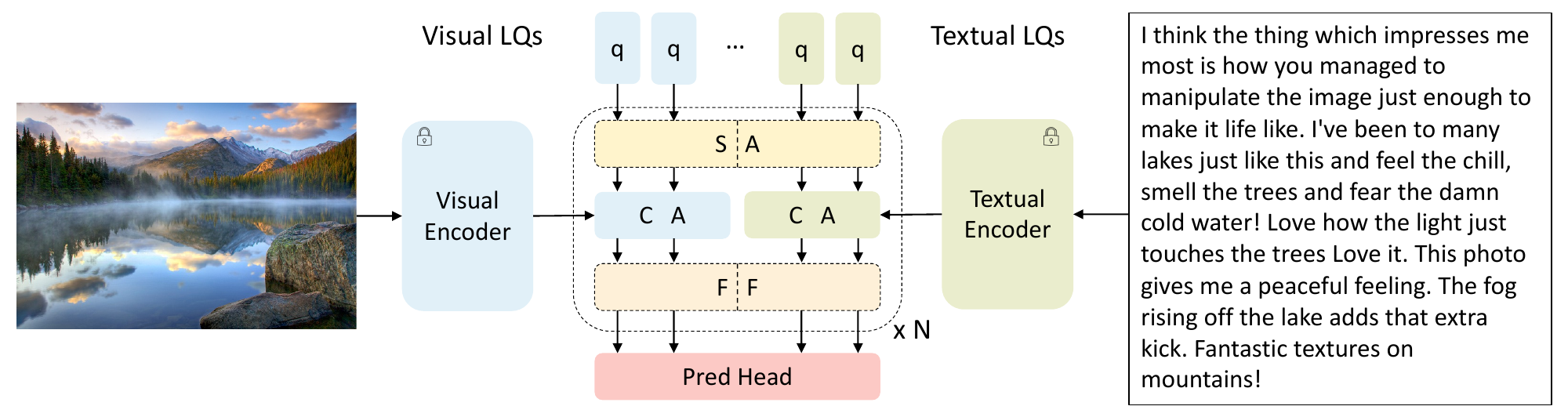}
\caption{The overall structure of the proposed multi-modal learnable queries (\methodname{}) method for IAA. The visual and textual LQs learn visual and textual aesthetic features through a replaceable self-attention layer, a separate multi-modal cross-attention layer with corresponding visual and textual pre-trained features, and a replaceable feed-forward layer. Finally, the averaged visual LQs and the averaged textual LQs are concatenated and fed into linear layers followed by Softmax to output the estimated aesthetic DOS of the input image.}
\label{fig:mmlq}
\end{figure*}

Recently, large-scale pre-trained models~\cite{radford2021learning, li2023blip} have demonstrated a strong capability in providing rich transferable knowledge for downstream tasks. Learnable queries and prompts~\cite{li2023blip, liu2023pre, ng2023soft} are shown to be effective ways to extract useful task-specific features from such pre-trained backbones for different modalities. Therefore, in this paper, we propose to utilize \underline{m}ulti-\underline{m}odal \underline{l}earnable \underline{q}ueries (\methodname{}) with large-scale frozen visual and textual encoders to extract multi-modal aesthetic features from both input images and their associated user comments. With a proper design for the multi-modal interaction blocks (MMIB) selected with comprehensive experiments, our proposed \methodname{} efficiently extracts and processes multi-modal aesthetic features and reaches new state-of-the-art performance on multi-modal IAA.

\section{Related Works}

Early works~\cite{datta2006studying, ke2006design} of IAA attempted to develop hand-crafted features corresponding to human-understandable photographic rules summarized by professional photographers. However, such photographic rules may not cover implicit aesthetic patterns that are difficult to summarize using human language. Consequently, later works~\cite{lu2014rapid, hosu2019effective, she2021hierarchical, he2022rethinking} proposed various deep learning models to capture such implicit aesthetic features showing promising results. Recently, as the prevalence of large-scale pretraining, frozen pre-trained image encoders~\cite{hou2022distilling, xiong2023federated, xiong2023image} have demonstrated an outstanding capability in providing rich transferable knowledge for IAA. Learnable queries~\cite{xiong2023image} is one efficient solution to extract relevant aesthetic features from such frozen image encoders.

With additional aesthetic guidance from user comments, a few recent works~\cite{zhu2023attribute, li2023image} attempted to extract multi-modal aesthetic features. Zhu \etal~\cite{zhu2023attribute} extracted multi-modal aesthetic features with attribute-based cross-modal attention and a gate unit. Li \etal\cite{li2023image} extracted attribute-grounded multi-modal aesthetic features with multi-modal memory network and cross-modal memory attention. However, their methods require additional pre-training on aesthetic attributes and suffer from performance degradation for images without dominant attributes. Unlike these attribute-based methods, we propose to utilize multi-modal learnable queries without attribute guidance to extract more diverse multi-modal aesthetic features from patch-level image features and token-level comment features with frozen visual and textual encoders.

\section{The Proposed Approach}

In this section, we present more details about our proposed \methodname{} method with an overall structure shown in Fig.~\ref{fig:mmlq}.

\subsection{Multi-modal Pre-trained Features}

For a given image $I$, a frozen pre-trained vision transformer (ViT) is employed to split the processed image into $N_p$ patch tokens and a prepend \texttt{[CLS]} token. The extracted pre-trained image features are a collection of visual token embeddings:
\begin{equation}
\vect{F}_v = \{\vect{v}_i\}_{i=1}^{N_p+1},
\end{equation}
where $\vect{v}_i \in \mathbb{R}^{H_v}$ with $H_v$ denoting the size of the visual features. Meanwhile, we accumulate all corresponding user comments on the given image $I$ into a long user comment $C$. A frozen pre-trained text encoder is employed to encode the comment into a collection of token-level embeddings:
\begin{equation}
\vect{F}_t = \{\vect{t}_j\}_{j=1}^{N_w},
\end{equation}
where $\vect{t}_j \in \mathbb{R}^{H_t}$ with $H_t$ denoting the size of the textual features. $N_w$ denotes the max comment length allowed including both word-level and special tokens at the beginning and end of the sentence (\ie, \texttt{[CLS]} and \texttt{[SEP]}). Comment lengths longer than $N_w$ are truncated to $N_w$ and comment lengths shorter than $N_w$ are filled with \texttt{[PAD]} tokens till $N_w$.

\subsection{Multi-modal Interaction Block}

Before the first multi-modal interaction block (MMIB), a set of learnable queries (LQs) are created to extract visual and textual aesthetic features, respectively. Specifically, they are a set of learnable embeddings that are shared among samples:
\begin{gather}
\vect{E}_v = \{\vect{e}_k\}_{k=1}^{N_v}, \vect{E}_t = \{\vect{e}_l\}_{l=1}^{N_t},
\end{gather}
where $\vect{e}_k, \vect{e}_l \in \mathbb{R}^{H_q}$ are visual and textual LQs with $H_q$ denoting the query embedding size. $N_v$ and $N_t$ denote the number of visual and textual LQs, respectively. The LQs are then normalized with separate layer normalization.

In each MMIB, both visual and textual LQs first go through a replaceable multi-head self-attention (SA) layer. The added SA layer could be shared among multi-modal LQs or separated based on modality, which can be expressed as:
\begin{gather}
\vect{Q}_{mh}, \vect{K}_{mh}, \vect{V}_{mh} =  \vect{E}_m\vect{W}_{mh}^Q, \vect{E}_m\vect{W}_{mh}^K, \vect{E}_m\vect{W}_{mh}^V, \\
\vect{A}_{mh} = \text{Softmax}(\frac{\vect{Q}_{mh}\vect{K}_{mh}^T}{\sqrt{d}})\vect{V}_{mh}, \label{eq:1} \\
\vect{E}_m = \text{LN}(\text{Concat}(\{\vect{A}_{mh}\}_h^{N_h})\vect{W}_m^O + \vect{E}_m), \label{eq:2}
\end{gather}
where $\vect{E}_m$ denotes either $\vect{E}_v$ or $\vect{E}_t$ in a separate SA layer while $\vect{E}_m = \text{Concat}(\vect{E}_v, \vect{E}_t) \in \mathbb{R}^{(N_v+N_t) \times H_q}$ in a shared SA layer. Weight matrices $\vect{W}_{mh}^Q, \vect{W}_{mh}^K, \vect{W}_{mh}^V \in \mathbb{R}^{H_q \times d}$ are employed to get down-sampled embeddings $\vect{Q}_{mh}, \vect{K}_{mh}, \vect{V}_{mh}$. $d$ is the hidden size of each head, and the scaling of $\frac{1}{\sqrt{d}}$ is to improve training stability. Finally, weight matrice $\vect{W}_m^O \in \mathbb{R}^{(N_h \times d) \times H_q}$ is employed to transform the concatenated multi-head attention results into the original dimension $H_q$ of $\vect{E}_m$, which is added into the original $\vect{E}_m$ followed by a layer normalization (LN) to get the new $\vect{E}_m$.

After the replaceable SA layer, the visual and textual LQs would extract aesthetic features from pre-trained visual and textual features respectively through a separate multi-head cross-attention (CA) layer. Different from the SA layer, keys and values are constructed from the pre-trained visual or textual features, which can be expressed as:
\begin{equation}
\vect{Q}_{mh}, \vect{K}_{mh}, \vect{V}_{mh} = \vect{E}_{m} \vect{W}_{mh}^Q, \vect{F}_{m} \vect{W}_{mh}^K, \vect{F}_m \vect{W}_{mh}^V,
\end{equation}
where $m$ denotes $v$ or $t$. Weight metrices $\vect{W}_{mh}^Q \in \mathbb{R}^{H_q \times d}$, $\vect{W}_{vh}^K, \vect{W}_{vh}^V \in \mathbb{R}^{H_v \times d}$, $\vect{W}_{th}^K, \vect{W}_{th}^V \in \mathbb{R}^{H_t \times d}$ are employed to transform LQs $\vect{E}_{m}$, visual features $\vect{F}_{v}$, textual features $\vect{F}_{t}$, respectively. Then, $\vect{Q}_{mh}, \vect{K}_{mh}, \vect{V}_{mh}$ go through a similar multi-head attention and an add \& norm procedure as illustrated in equations~\ref{eq:1},~\ref{eq:2} to obtain the new $\vect{E}_{m}$.

Followed by the replaceable SA layer and the separate CA layer, a replaceable feed-forward (FF) layer is added to further process the LQs. It comprises two linear layers, a GELU activation~\cite{hendrycks2016gaussian}, and an add \& norm procedure as in the SA and CA layers, which can be expressed as:
\begin{equation}
\vect{E}_m = \text{LN}(\text{FC}(\text{GELU}(\text{FC}(\vect{E}_m)))+\vect{E}_m),
\end{equation}
where the LQs are projected to a larger intermediate dimension $H_i = 4 \times H_q$, and then back to $H_q$. Similar to the SA layer, the added FF layer could be shared among multi-modal LQs or separated based on modality.

\subsection{Prediction Head for IAA}

After $N$ MMIBs, both visual and textual LQs accumulate adequate aesthetic patterns for the final prediction. We take the average of LQs within each modality and concatenate the two averaged LQs. One linear layer (output dimension of $H_q$) with GELU activation and another linear layer (output dimension of 10) with Softmax activation are employed on the concatenated LQs to output the final prediction:
\begin{gather}
\vect{E}_a = \text{Concat}(\Bar{\vect{E}}_v, \Bar{\vect{E}}_t), \\
\hat{\vect{D}}_a = \text{Softmax}(\text{FC}(\text{GELU}(\text{FC}(\vect{E}_a)))),
\end{gather}
where $\Bar{\vect{E}}_v, \Bar{\vect{E}}_t \in \mathbb{R}^{H_q}$ denote the averaged visual and textual LQs, respectively. $\vect{E}_a \in \mathbb{R}^{2 H_q}$ denotes the concatenated multi-modal aesthetic features. $\hat{\vect{D}_a} \in \mathbb{R}^K$ denotes the predicted K-scale aesthetic DOS. Following previous works~\cite{she2021hierarchical, zhu2023attribute, li2023image}, we adopt Earth Mover’s Distance (EMD) loss to optimize the prediction:
\begin{equation}
\mathcal{L}(\vect{D}_a, \hat{\vect{D}}_a) = \sqrt{\frac{1}{K}\sum_{k=1}^K \lvert CDF_{\vect{D}_a}(k)-CDF_{\hat{\vect{D}}_a}(k) \rvert ^2},
\end{equation}
where $CDF_{\vect{D}_a}$ and $CDF_{\hat{\vect{D}}_a}$ are cumulative density functions for the ground truth and predicted aesthetic DOS, respectively.

\section{Experimental Evaluation}

\begin{figure}[!t]
\centering
\includegraphics[width=1\linewidth]{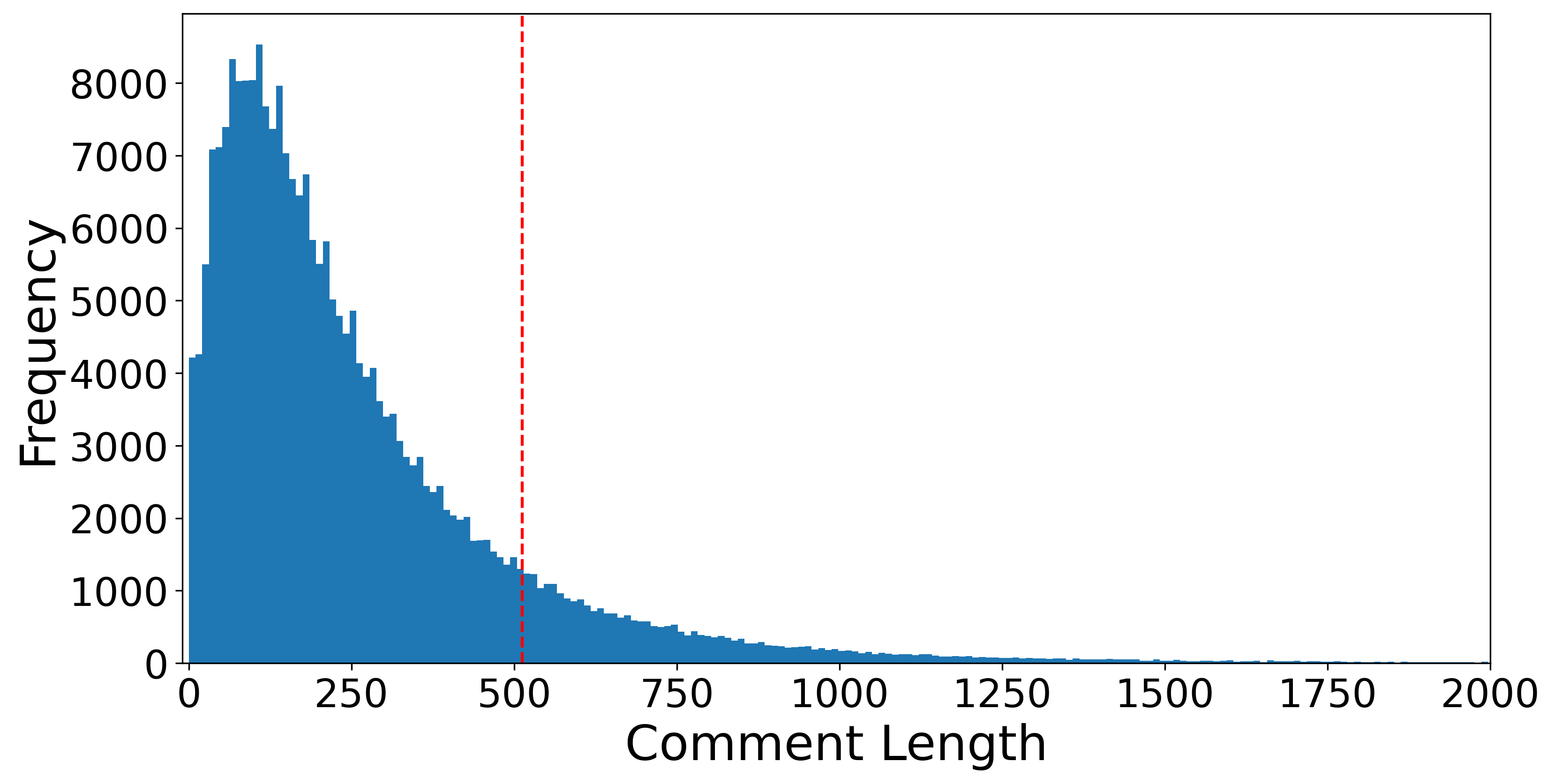}
\caption{The histogram of the accumulated comment lengths for each image in AVA-Comments~\cite{zhou2016joint}. The vertical dotted line in red indicates a comment length of 512, which is the maximum comment length ($N_w$) allowed in our experiments.} 
\label{fig:histogram}
\end{figure}

\subsection{Dataset}

The dataset we use for experiments is the benchmark IAA dataset, the AVA dataset~\cite{murray2012ava}. It contains more than 250,000 images downloaded from the DPChallenge website, where each image received 78 to 549 aesthetic scores (average of 210) on a scale of 1 to 10, where 1 and 10 denote the lowest and highest aesthetics, respectively. The AVA-Comments~\cite{zhou2016joint} dataset provides the corresponding user comments on the images in AVA, which were crawled from the original links of the images with a histogram of accumulated comment length for each image as in Fig~\ref{fig:histogram}. We use the same split as in~\cite{she2021hierarchical}, where 235,510 images are used for training, and 19,998 images are used for testing.

\begin{table*}[!t]
\caption{Comparison with the state-of-the-art methods on the train-test split from~\cite{she2021hierarchical}.\label{tab:peer_comparison}}
\centering
\begin{tabularx}{\textwidth}{XXXccccc}
\toprule
Methods & Modality & Input size & SRCC$\uparrow$ & PLCC$\uparrow$ & Acc(\%)$\uparrow$ & MSE$\downarrow$ & EMD$\downarrow$ \\
\midrule
NIMA~\cite{talebi2018nima} & Image & 224$\times$224 & 0.612 & 0.636 & 81.51 & - & 0.050 \\
AFDC~\cite{chen2020adaptive} & Image & 320$\times$320 & 0.649 & 0.671 & 83.24 & 0.271 & 0.045 \\
HLA-GCN~\cite{she2021hierarchical} & Image & 300$\times$300 & 0.665 & 0.687 & 84.60 & 0.255 & 0.043 \\
KD-IAA~\cite{hou2022distilling} & Image & Full resolution & 0.732 & 0.751 & 85.30 & - & - \\
TCD-IAA~\cite{hou2023towards} & Image & Full resolution & 0.767 & 0.783 & 86.20 & - & - \\
\midrule
AAM-Net~\cite{zhu2023attribute} & Multi-modal & - & 0.780 & 0.803 & 85.69 & 0.176 & 0.037 \\
AMM-Net~\cite{li2023image} & Multi-modal & - & 0.816 & 0.830 & 87.58 & 0.160 & 0.035 \\
\midrule
\methodname{} (CLIP) & Multi-modal & 224$\times$224 & \textbf{0.876} & \textbf{0.896} & \textbf{89.03} & \textbf{0.094} & \textbf{0.029} \\
\methodname{} (EVA-CLIP) & Multi-modal & 224$\times$224 & \textbf{0.879} & \textbf{0.899} & \textbf{88.99} & \textbf{0.092} & \textbf{0.028} \\
\bottomrule
\end{tabularx}
\end{table*}

\subsection{Experiment Settings}

Following~\cite{li2023blip, xiong2023image}, we set the query hidden size $H_q$ to 768, and the default number of visual and textual queries $N_v, N_t$ to 2. We use $N_h=12$ heads in multi-head attention layers, yielding a hidden size $d = 64$ for each head. For input images, we only apply horizontal flipping with $p = 0.5$ during training and directly resize the image to $224 \times 224$. Frozen pre-trained ViTs with $14 \times 14$ patch sizes are applied as the visual encoder, resulting in $N_p = 16 \times 16 = 256$ patches for each input image. For input user comments, the frozen text encoder we adopt is BERT$_\text{base}$~\cite{devlin2018bert} which outputs token-level textual features with $H_t = 768$. The max comment length $N_w$ is set to the maximum length allowed in BERT$_\text{base}$ (\ie, 512), approximately the 90th percentile of the histogram in Fig~\ref{fig:histogram}. We train \methodname{} with a batch size of 128 for 10 epochs with the Adam optimizer. The learning rate is initially set to $3 \times 10^{-5}$, and multiplied by 0.1 every two epochs.

\subsection{Comparison with State-of-the-art Methods}

\begin{table}[!t]
\caption{Performance with different MMIB design. \xmark, \cmark, \cmark\cmark denote no, shared, and separate layers, respectively. \label{tab:design}}
\centering
\begin{tabular}{ccccc}
\toprule
SA & FF & SRCC$\uparrow$ & PLCC$\uparrow$ & Acc(\%)$\uparrow$ \\
\midrule
\xmark & \xmark & 0.874 & 0.894 & 88.92\\
\xmark & \cmark & \textbf{0.876} & \textbf{0.896} & \textbf{89.03} \\
\xmark & \cmark\cmark & \textbf{0.876} & \textbf{0.896} & 88.96 \\
\midrule
\cmark & \xmark & 0.875 & 0.894 & 88.88 \\
\cmark & \cmark & 0.874 & 0.894 & 88.72 \\
\cmark & \cmark\cmark & 0.873 & 0.893 & 88.73 \\ %
\midrule
\cmark\cmark & \xmark & 0.873 & 0.892 & 88.78 \\
\cmark\cmark & \cmark & 0.873 & 0.893 & 88.81 \\ %
\cmark\cmark & \cmark\cmark & 0.875 & 0.894 & 88.77\\
\bottomrule
\end{tabular}
\end{table}

We compare the performance of \methodname{} with previous image-based and multi-modal methods in Table~\ref{tab:peer_comparison}. Following previous works~\cite{she2021hierarchical, zhu2023attribute, li2023image}, we evaluate \methodname{} with binary accuracy (Acc) for binary aesthetic classification (where a MOS of 5 is the boundary); with Spearman’s rank correlation coefficient (SRCC), Pearson linear correlation coefficient (PLCC), Mean Squared Error (MSE) for MOS regression; with EMD loss for DOS prediction. In general, multi-modal methods have higher performance compared to image-based methods, due to the additional aesthetic knowledge in the user comments. We report the results of \methodname{} with ViT-L/14 from CLIP~\cite{radford2021learning} and ViT-G/14 from EVA-CLIP~\cite{fang2023eva}, where the former is the default vision encoder in our experiments with a $H_v$ of 1024, and the latter has a $H_v$ of 1408. With multi-modal learnable queries to extract multi-modal aesthetic features, our method reaches state-of-the-art performance with both vision encoders on all the metrics with a low input image resolution (\ie, $224 \times 224$).
The performance of \methodname{} (EVA-CLIP) beats previous methods by 7.7\% and 8.3\% in terms of SRCC and PLCC, respectively.

\subsection{Ablation Studies}

\noindent\textbf{Effects of different MMIB design}: We first explore how \methodname{}'s performance varies with different MMIB designs. Specifically, we set $N$ of MMIBs to 6, and focus on the designing of the self-attention (SA) and feed-forward (FF) layers. Table~\ref{tab:design} shows the performance with no / shared / separate SA layer paired with no / shared / separate FF layer. Overall, MMIB with different designs results in similar performance. It shows the main contribution of MMIB comes from the cross-attention (CA) between multi-modal LQs and their corresponding multi-modal pre-trained features. The best performance occurs when shared FF layers are added after the CA layer of each modality. The inferior performance with shared / separate SA layers could be caused by the feature space gap among LQs of different modalities and over-emphasis on the difference among LQs of the same modality, respectively.

\begin{table}[!t]
\caption{Performance with different modalities. \label{tab:modality}}
\centering
\begin{tabular}{ccccc}
\toprule
Visual & Textual & SRCC$\uparrow$ & PLCC$\uparrow$ & Acc(\%)$\uparrow$\\
\midrule
\cmark & \xmark & 0.725 & 0.743 & 85.25\\ 
\xmark & \cmark & 0.822 & 0.852 & 85.90 \\
\midrule
\cmark & \cmark & \textbf{0.876} & \textbf{0.896} & \textbf{89.03} \\
\bottomrule
\end{tabular}
\end{table}

\noindent\textbf{Effects of different modality}:
To explore the contributions from each modality, we keep the number of LQs (\ie, $N_v$ or $N_t$) for one modality unchanged, and remove the LQs, encoder, and cross-attention layer corresponding to the other modality. As shown in Table~\ref{tab:modality}, the performance with multi-modality is superior to single visual or textual modality, which reflects the complementarity of the two modalities in extracting aesthetic features. Interestingly, the performance of using textual modality only is superior to using visual modality only. It could be attributed to the rich sentiment clues available in user comments that are beneficial for IAA.

\begin{figure}[!t]
\centering
\includegraphics[width=1\linewidth]{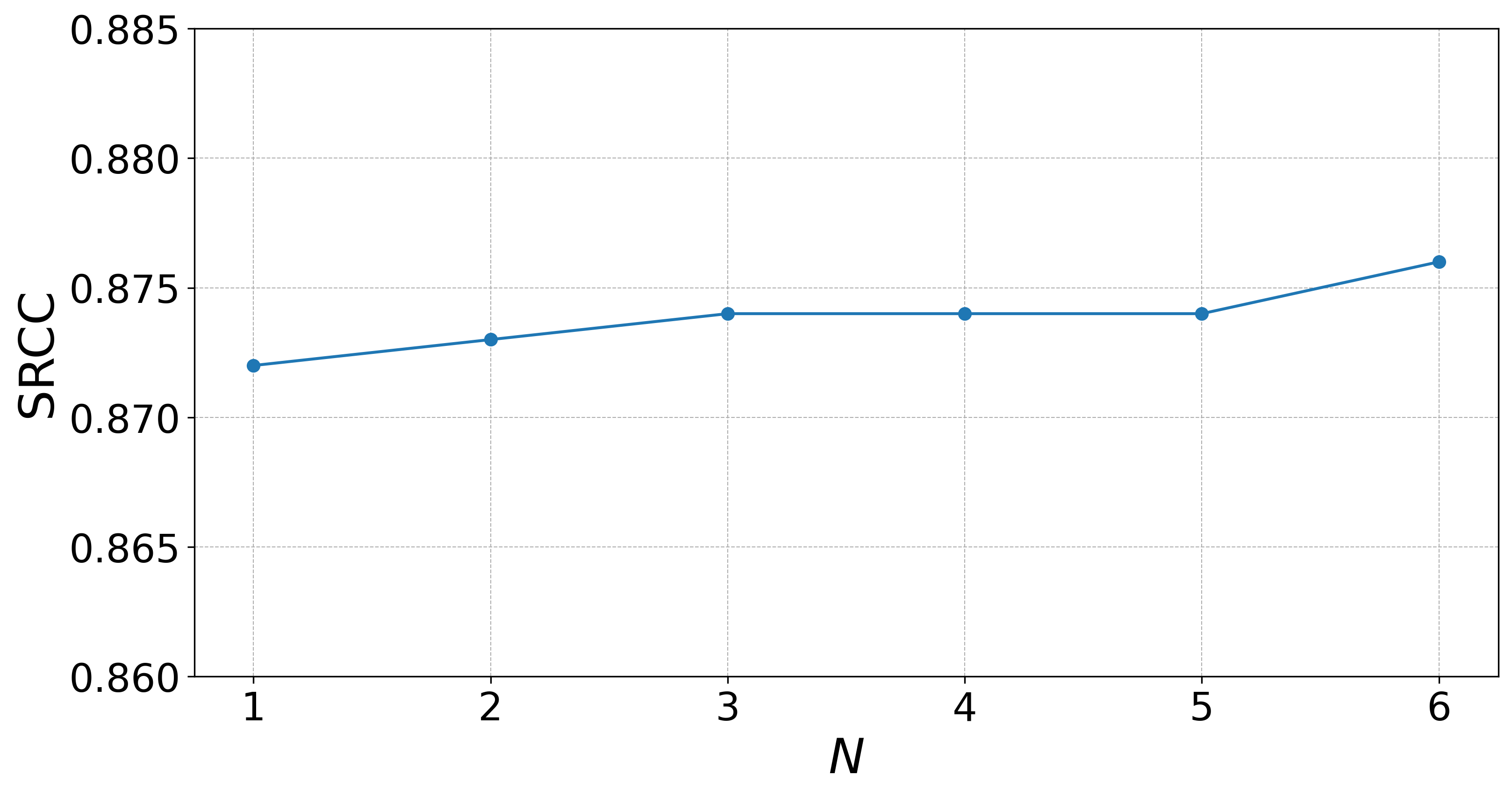}
\caption{Performance with different numbers of MMIBs.}
\label{fig:num_layers}
\end{figure}

\begin{figure}[!t]
\centering
\includegraphics[width=1\linewidth]{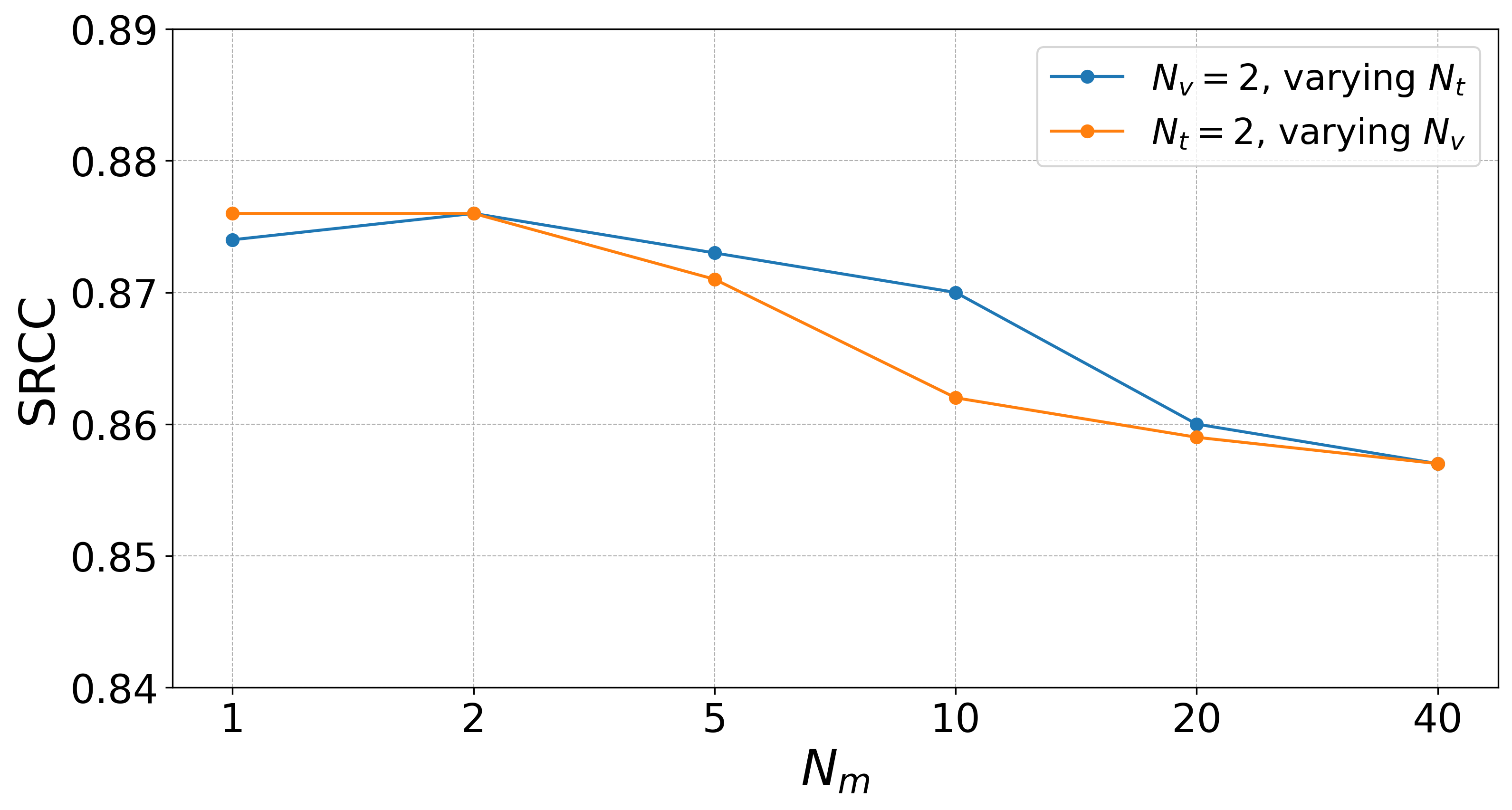}
\caption{Performance with different numbers of LQs.}
\label{fig:num_queries}
\end{figure}

\noindent\textbf{Effects of different model complexity}:
Since model performance is often directly affected by model complexity, we compare model performance with different model complexity. Specifically, we explore the effect of different numbers of MMIBs in Fig.~\ref{fig:num_layers}, and different numbers of LQs in Fig.~\ref{fig:num_queries}. Surprisingly, involving more LQs tends to degrade the performance of \methodname{}. It could be due to the additional knowledge captured by additional LQs being redundant for image aesthetics. On the contrary, the performance of \methodname{} is fairly stable with varying $N$. These observations demonstrate the effectiveness of a limited number of MMIBs and LQs in capturing adequate visual and textual aesthetic features.

\subsection{Example Results}

\begin{figure}[!t]
\centering
\includegraphics[width=1\linewidth]{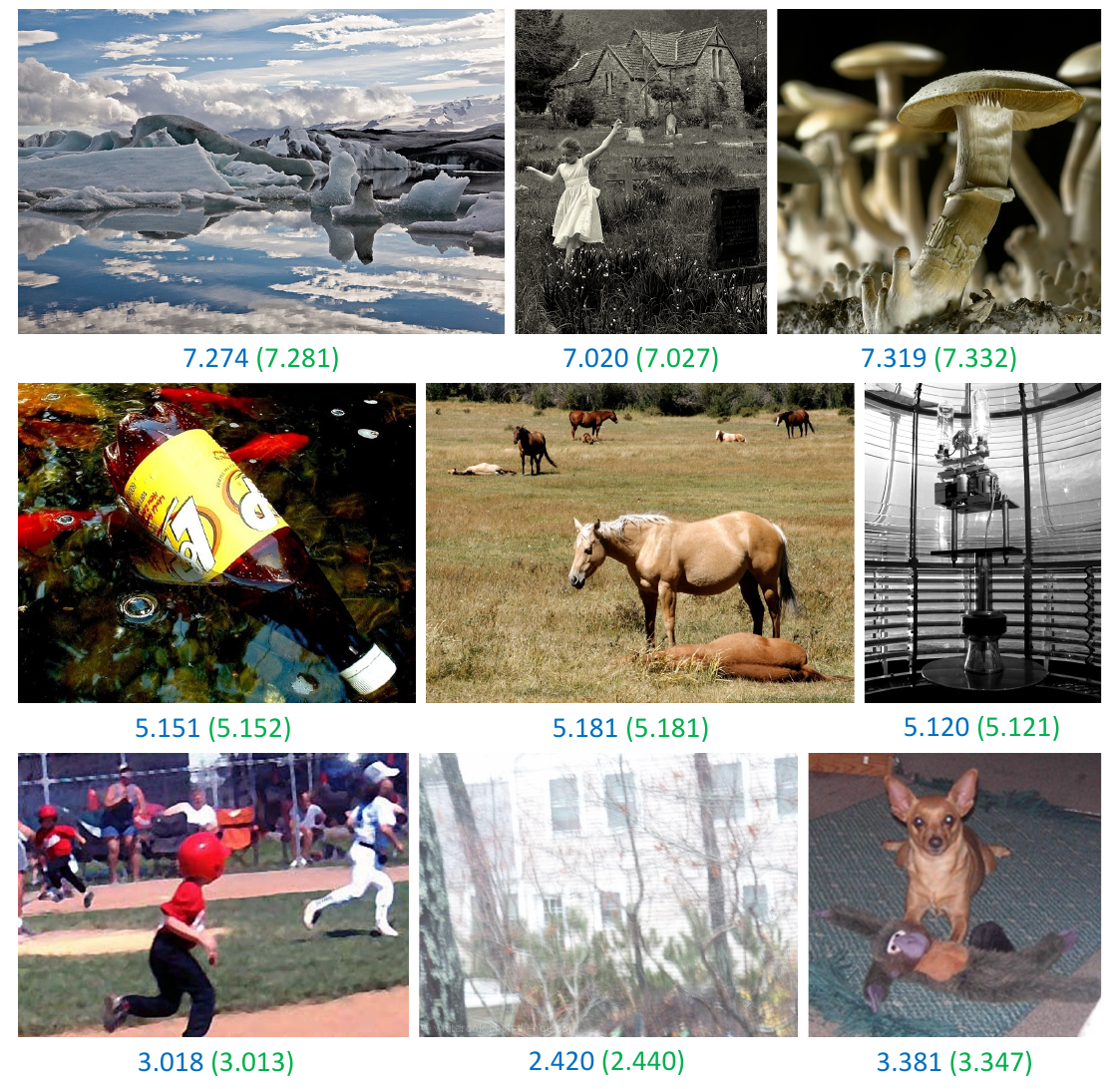}
\caption{Example results on images from the test set of the AVA dataset with \methodname{}. The blue and (green) numbers below each image indicate their corresponding predicted and (ground truth) aesthetic MOSs, respectively.}
\label{fig:pos_examples}
\end{figure}

Several example results of \methodname{} (EVA-CLIP) on images from the test set of AVA are shown in Fig~\ref{fig:pos_examples}. The examples cover a wide range of aesthetic levels (\ie, relatively high, moderate, and low ground truth MOSs from top row to bottom row) and visual content (\eg, landscape, humans, animals, plants, objects, \etc). The small errors between their predicted and ground truth MOSs demonstrate the strong capability of \methodname{} in IAA for images with various aesthetic levels and visual content.

\begin{table}[!t]
\caption{Performance when different portions of user comments are available during inference. \label{tab:limitation}}
\centering
\begin{tabular}{ccccc}
\toprule
Comment(\%) & SRCC$\uparrow$ & PLCC$\uparrow$ & Acc(\%)$\uparrow$ \\
\midrule
100 & 0.879 & 0.899 & 88.99 \\ 
60 & 0.866 & 0.888 & 87.71 \\
20 & 0.818 & 0.846 & 83.86 \\
\bottomrule
\end{tabular}
\end{table}

\subsection{Limitation} \label{sec:limitation}

Although multi-modal IAA methods~\cite{zhu2023attribute, li2023image} like \methodname{} demonstrate extraordinary performance in IAA, they share a common practical limitation of unpredictable performance when only a limited portion of user comments are available. In Table~\ref{tab:limitation}, we compare the performance of \methodname{} (EVA-CLIP) when different portions of user comments are available during inference. It can be observed that the performance of \methodname{} does degrade as fewer user comments are available. It could be due to the mismatch in sentiment between the randomly selected comments and all comments which is critical in IAA because everyone's opinion matters. Nevertheless, with only 20\% of the user comments during inference, the performance of \methodname{} still outperforms previous multi-modal IAA methods in terms of SRCC and PLCC, indicating the effectiveness of \methodname{} in capturing multi-modal aesthetic details for IAA.

\section{Conlusion and Future Work}

In the paper, we propose the multi-modal learnable queries (\methodname{}) method, which adopts learnable queries with frozen image and text encoders to efficiently extract multi-modal aesthetic features from input images and their associated user comments. Extensive experiments have demonstrated that our proposed \methodname{} method largely outperforms existing state-of-the-art IAA methods on the benchmark AVA dataset with user comments provided by the AVA-Comments dataset.

In future research, we plan to tackle the limitations as mentioned in Section~\ref{sec:limitation} by incorporating additional submodules based on aesthetic-aware image-to-text retrieval or image captioning, both of which no longer rely on explicit corresponding user comments during inference.

\bibliographystyle{IEEEtran}
\bibliography{main}

\end{document}